\documentclass{ceurart}
\usepackage{xcolor}

\usepackage{url}
\usepackage{booktabs}
\usepackage{array}
\usepackage{graphicx}
\usepackage{hyphenat}
\usepackage{pifont}

\begin{document}

\copyrightyear{2026}
\copyrightclause{Copyright for this paper by its authors. Use permitted under Creative Commons License Attribution 4.0 International (CC BY 4.0).}
\conference{ISWC 2026 Companion Volume, October 25--29, 2026, Bari, Italy}

\title{FedV-KGQA in Practice: Design Lessons and an Interactive Prototype}

\author{Md Saikat Islam Khan Bappy}[%
  orcid=0009-0009-1768-6102,
  email=islamm9@rpi.edu
]

\author{Oshani Seneviratne}[%
  orcid=0000-0001-8518-917X,
  email=senevo@rpi.edu
]

\address{Rensselaer Polytechnic Institute,
Troy, NY 12180, USA}
\begin{abstract}
Knowledge graph question answering usually assumes that one system can reach the
whole graph. In practice, facts are often held by organizations that share
entity identifiers but own disjoint relation types, so no single party sees a
complete reasoning chain. This poster presents the empirical findings of
FedV-KGQA on multi-hop question answering
over such vertically partitioned graphs. Each silo enriches its local graph and
trains a knowledge graph embedding on its own triples. A server then
concatenates the silo-specific entity views, anchors the projected question at
the topic entity, and ranks candidates by similarity. Raw triples and relation
embeddings never leave a silo. Comparing the FedV-KGQA experiments with one another yields three results. First, federated fusion recovers most
of the centralized accuracy, while a single silo recovers little. Second,
anchoring and enrichment matter more than the choice of embedding model. Third,
the cheapest encoder depends on the target accuracy rather than on parameter
count. This poster paper contributes that cross-experiment comparison, four design lessons drawn from it, and an interactive prototype that runs real
inference and traces the full pipeline, per question, on released checkpoints.
\end{abstract}


\maketitle

\section{Introduction}

Multi-hop knowledge graph question answering (KGQA) chains several facts to
reach an answer entity: answering ``which actors starred in films directed by
Christopher Nolan?'' takes one hop to the films and a second to their cast.
Almost all KGQA systems assume a single system can query the whole knowledge
graph (KG) to follow such a chain~\cite{saxena2020improving,jiang2023unikgqa},
and recent methods that add large language models keep that assumption, since
the evidence is gathered in one place~\cite{luo2025bridging,ma2025large}.
That assumption fails when the hops reside in different organizations: a film studio may hold
\texttt{director} relations, a streaming platform \texttt{cast} relations, and a metadata service
\texttt{genre} relations, so each party owns a slice of the same entities and none
can trace a chain across the slices, while governance, sovereignty, and
commercial intellectual property constraints usually forbid merging the triples. This is a
\emph{vertical} partition of the KG, in which parties share entities as samples
but hold disjoint relations as feature views~\cite{khan2024fed, tran2024differentially}. Existing
federated knowledge graph embedding (KGE) work instead assumes horizontal
partitions, where clients share the relation vocabulary but hold different
triples, and targets link prediction rather than multi-hop
QA~\cite{chen2021fede,hu2025learning,zhu2025parameter}. Federated SPARQL engines answer queries across autonomous sources, and
recent work continues to advance source selection and scale to large
federations~\cite{ogura2025efficient,aimonier2024fedup}. These engines require
queryable endpoints and a structured query, whereas our silos expose only
embeddings and run no query over their triples. Neither line covers a vertical
split answered from natural language, which is where FedV-KGQA sits, ranking
answers across relation silos while keeping triples and relation parameters
inside each silo~\cite{bappy2026fedv}.

FedV-KGQA~\cite{bappy2026fedv} answers six research questions in separate sections but never compares
them, and that comparison is what this poster adds: model choice looks decisive
until the ablation shows anchoring and enrichment matter more; silo count is a
communication parameter in one section and a representation-quality parameter in
another; and encoder choice looks settled by accuracy until the communication
experiment shows the cheapest encoder depends on the target. Section~\ref{sec:lessons} turns that comparison into four design lessons, and an
interactive prototype (Section~\ref{sec:prototype}) runs live inference on
released checkpoints, tracing a single question through every stage of the
pipeline, a per-question view FedV-KGQA~\cite{bappy2026fedv} does not provide.
 
\section{Framework Overview}

\begin{figure}[t]
  \centering
\includegraphics[width=0.98\linewidth]{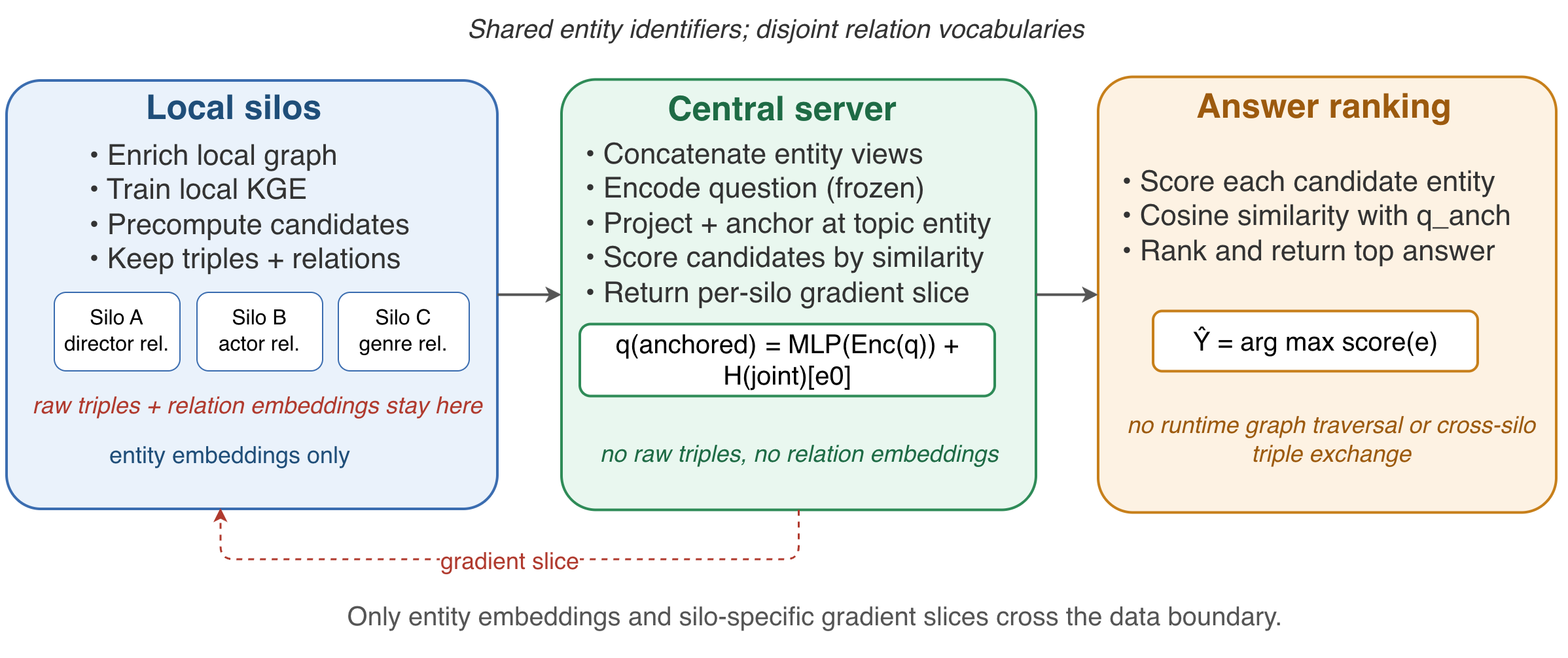}
  \caption{Workflow of FedV-KGQA. Only entity embeddings and silo-specific gradient slices
cross the boundary.}
  \label{fig:workflow}
  \vspace{-10pt}
\end{figure}

Figure~\ref{fig:workflow} shows the four stages: local enrichment, local KGE
training, server-side fusion with QA training, and inference. Each silo first
enriches its local graph with inverse-property and property-chain rules, without
any triple leaving the silo, which improves entity representation quality and
answer reachability. The inverse rules matter more than they look, because under
a vertical split an answer entity often appears only as a tail in its own silo,
so its embedding is trained from one role only and gives a weak similarity
signal; adding the inverse triple restores that signal. Each silo also
precomputes a topic-conditioned candidate set by bounded neighborhood expansion,
then trains a local KGE model on its enriched triples, for which we test
TransE~\cite{bordes2013translating}, DistMult~\cite{yang2014embedding},
ComplEx~\cite{trouillon2016complex}, and RotatE~\cite{sun2019rotate}. Raw
triples and relation embeddings stay local, and only the entity embedding matrix
is shared.

The server concatenates the silo-specific entity embeddings into a joint matrix
$\mathbf{H}_{\mathrm{joint}}$ instead of averaging them, which preserves the
geometric view of each silo. A frozen encoder maps the question to a vector, a
multilayer perceptron (MLP) projects it into the joint entity space, and the question
is anchored at the topic entity $e_0$,
\begin{equation}
  \mathbf{q}_{\mathrm{anch}} = \mathrm{MLP}(\mathrm{Enc}(q)) + \mathbf{H}_{\mathrm{joint}}[e_0],
\end{equation}
which grounds ranking in the relevant neighborhood. The topic entity $e_0$
is supplied with each question by the benchmarks and is not predicted, so
entity linking lies outside the scope of this evaluation. Candidates are
scored by cosine similarity with $\mathbf{q}_{\mathrm{anch}}$. 
The MLP is the only trainable component. Given a question whose gold answers
form the set $\mathcal{A}$, QA training minimizes a margin ranking loss over
the precomputed candidate set,
\begin{equation}
  \mathcal{L}^{\mathrm{QA}} = \max\!\Big(0,\; \gamma
  + \!\!\max_{e^-\in\,\mathcal{C}(e_0)\setminus\mathcal{A}}\!\!\mathrm{score}(e^-)
  \;-\; \max_{e^+\in\,\mathcal{A}}\mathrm{score}(e^+)\Big),
\end{equation}
which penalizes the model whenever the hardest negative candidate scores close
to the best gold answer. Because the encoder is frozen, the projection head
alone learns to place question vectors near the answer region of the joint
space, while relation embeddings receive no gradient and stay at their local
values. The gradient with respect to the concatenated entity matrix decomposes
into silo-specific column slices, so each slice goes back only to its own silo. At
inference the server reuses the fine-tuned entity matrices and the precomputed
candidates, without runtime graph traversal or cross-silo triple exchange.

\paragraph{Scope.} The method answers path-shaped questions. Each
question comes with a topic entity, and the answer must be an entity that the
bounded expansion reaches. Questions that need aggregation, comparison,
counting, or time constraints fall outside a ranking formulation, as do
questions whose answer is not an entity. Accuracy also depends on how much
structure each silo holds and on how well entity identifiers align across
silos.

\section{Experimental Design}
\label{sec:design}

We use three benchmarks of differing scale and domain: MetaQA (43,000+ movie
entities, one- to three-hop questions)~\cite{zhang2018variational}, PathQuestion
(person-centric two- and three-hop chains from
Freebase)~\cite{zhou2018interpretable}, and WebQuestionsSP (WebQSP; 985,000+
entities, open-domain, up to two hops)~\cite{yih2016value}. Relations are
assigned by semantic category to three, five, or seven silos with shared entity
identifiers, so each relation belongs to exactly one silo and the silos jointly
cover the full vocabulary. Twelve configurations pair four KGE models with a
frozen BERT~\cite{devlin2019bert}, DistilBERT~\cite{sanh2019distilbert}, or
RoBERTa~\cite{liu2019roberta} encoder, evaluated by mean reciprocal rank (MRR)
and Hits@$K$. The experiments span two- and three-hop accuracy, ablations of
anchoring and enrichment, adapted federated and centralized baselines, Gaussian
perturbation of transmitted entity embeddings, and communication costs required to reach a target Hits@3, with three silos as the default partition.

Partitions differ substantially in what each silo can learn. Under the
three-silo split, and counting inverse relations, MetaQA silos hold 4, 3, and 8
relations over 63K, 129K, and 84K enriched triples; PathQuestion silos hold 3,
5, and 5 relations over 18K, 168K, and 191K triples; and WebQSP silos hold 332,
847, and 2{,}098 relations over 0.59M, 0.99M, and 1.89M triples. Silo count and
per-silo structure therefore vary together, which is why silo count alone does
not predict accuracy. The \emph{local-only} baseline used below removes fusion
entirely: each silo trains, expands candidates, and ranks using only its own
embeddings and its own triples, and we report the best single silo. Built from
one silo alone, the candidate set covers at most 54\%, 35\%, and 46\% of gold
answers on MetaQA, PathQuestion, and WebQSP, against 99\%, 100\%, and 78\% when
the silos are combined. Coverage caps
attainable accuracy, so WebQSP's results are measured against a ceiling of
0.78.
 

\section{Empirical Findings}
\label{sec:result}

\begin{figure}[t]
  \centering
\includegraphics[width=0.70\linewidth]{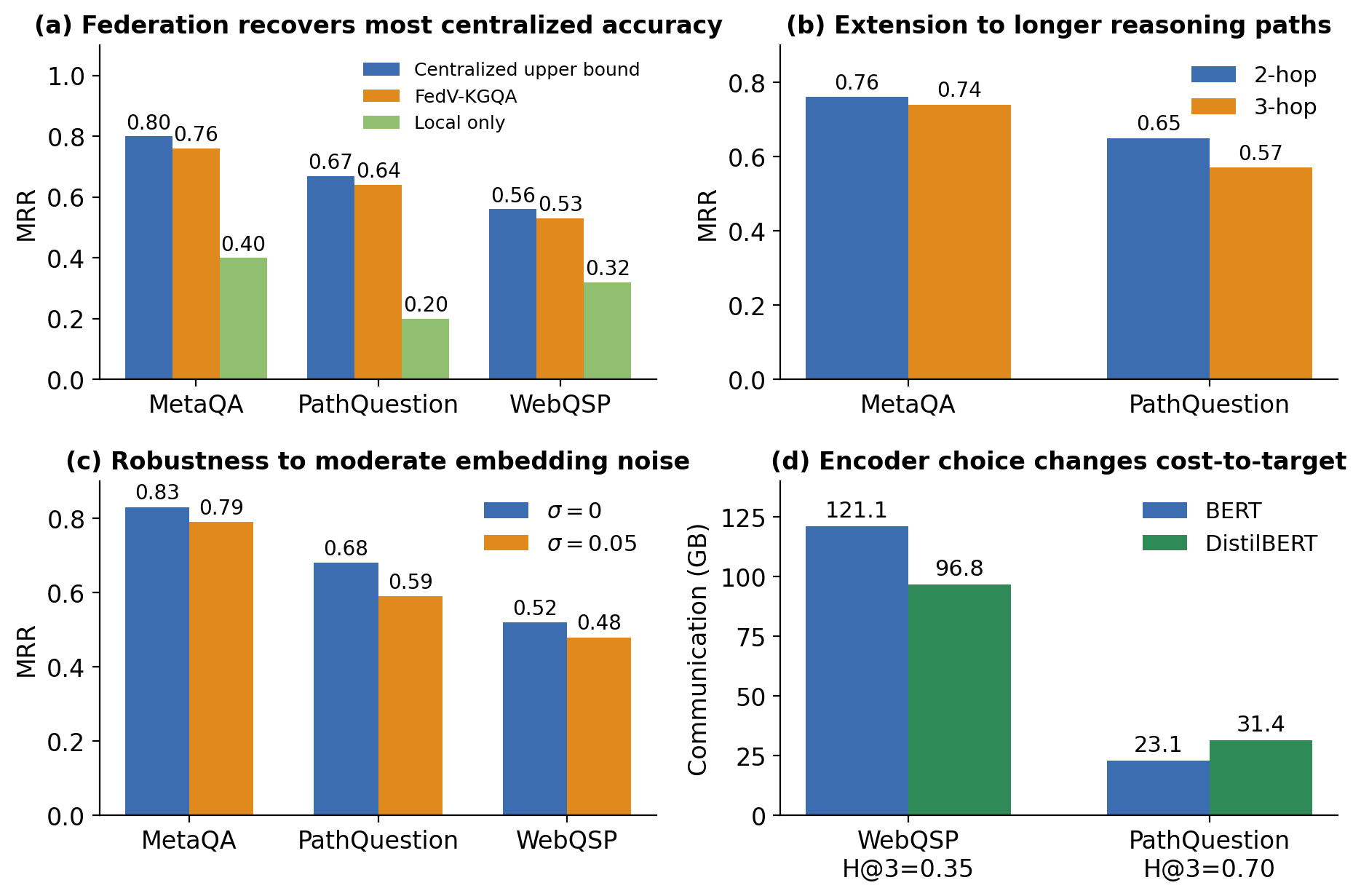}
  \caption{Cross-experiment synthesis. (a) DistilBERT+TransE baseline comparison, three
  silos. (b) BERT+TransE from two to three hops, three silos. (c) BERT+TransE
  robustness, five silos. (d) Communication for selected Hits@3 targets.}
  \label{fig:synthesis}
\end{figure}

\textbf{Finding 1: Fusion works, and the model matters less than the mechanism.}
With DistilBERT+TransE, FedV-KGQA trails a centralized upper bound (Figure~\ref{fig:synthesis} (a)) by 0.04 MRR on
MetaQA and 0.03 on PathQuestion and WebQSP, whereas the best single-silo baseline drops to
0.40, 0.20, and 0.32. No single partition holds the full chain. Yet concatenated
entity views recover most of what centralized access gives, reaching 0.76, 0.65,
and 0.54 MRR and beating adapted FedE~\cite{chen2021fede} by up to 0.08 and
adapted RelChain~\cite{jin2023improving} by up to 0.12. The KGE model is
secondary: TransE is most stable (0.54 MRR on WebQSP with BERT, against 0.41 for
DistMult, 0.48 for ComplEx, 0.49 for RotatE), but on MetaQA all four range from 0.71 to 0.76. BERT and DistilBERT stay close despite DistilBERT's smaller size,
while RoBERTa trails throughout, suggesting its \texttt{[CLS]} vectors align
poorly with the KGE spaces. We evaluate translational and bilinear scoring functions only, since the
question here is how fusion and anchoring behave under a vertical split rather
than which scoring function is strongest. Message-passing encoders, such as graph neural networks (GNNs), are excluded for a structural reason: each silo would aggregate over a graph missing most of
its edges by construction, so the neighborhood a GNN exploits is precisely what
the partition removes. Whether enrichment restores enough structure to make
aggregation worthwhile is left to future work.

\textbf{Finding 2: Partitioning and path length behave non-uniformly.}
Silo count is not a simple knob. With BERT+TransE, WebQSP falls only from 0.54 to
0.51 MRR from three to seven silos and MetaQA peaks at 0.83 with five, while
PathQuestion rises to 0.68 at five silos then drops to 0.55 at seven. So moderate
partitioning can aid specialization while heavy fragmentation weakens embeddings,
and relation placement shapes representation quality as much as silo count
does, as the partition statistics in Section~\ref{sec:design} indicate.
Longer paths need no architectural change (Figure~\ref{fig:synthesis} (b)). On MetaQA, BERT+TransE falls only from 0.76
to 0.74 MRR (0.93 Hits@10 held), but on PathQuestion it falls 0.65 to 0.57 and
0.96 to 0.90, tracking its smaller, more heterogeneous training set. So
longer-path generalization should not be inferred from one benchmark.

\textbf{Finding 3: Both components help, and cost is target-dependent.}
Removing anchoring lowers MRR from 0.64 to 0.57 on PathQuestion and 0.53 to 0.47 on
WebQSP, while removing enrichment causes larger drops on WebQSP (0.19 MRR, 0.28 Hits@10).
Enrichment strengthens local structure and candidate reachability, anchoring
focuses ranking on the right neighborhood. Robustness holds at low noise (Figure~\ref{fig:synthesis} (c)): at
$\sigma=0.05$, MRR drops only to 0.79, 0.59, and 0.48, and decline steepens
above $\sigma=0.10$. This is a robustness analysis, not a privacy
guarantee~\cite{peng2021differentially}. Total communication (Figure~\ref{fig:synthesis} (d)) is
$C_{\mathrm{total}} = 2TK|\mathcal{E}|d \cdot 4$ bytes, so cost is
target-dependent. DistilBERT reaches Hits@3 of 0.35 on WebQSP for 96.8 vs 121.1
GB, but BERT is cheaper at higher targets (23.1 vs 31.4 GB on PathQuestion at
Hits@3 of 0.70), making encoder efficiency a cost-to-target question~\cite{zhu2025parameter}. Cost scales with the entity vocabulary, not the triple count, so sparse updates restricted to the candidate sets should reduce it substantially. We have not measured this. 

\section{Interactive Prototype}
\label{sec:prototype}

The prototype turns the pipeline into something a viewer can drive. It loads the
released checkpoints (TransE with BERT, three silos, two-hop MetaQA) and runs
real inference. The user picks a question, and each stage streams in turn: the topic entity, the private triple each silo holds, two-hop candidate
filtering, fusion, and the ranked answers with per-question MRR and Hits. The silo
triples are displayed for demonstration only, to show what each silo holds
and how the answer is assembled across silos; in deployment, neither the
user nor the server sees them. The same trace exposes properties that aggregate metrics hide. First, the
\emph{cross-silo dependency} is visible per query. One can see which silo
supplies each hop and confirm that no silo holds the whole chain. Second, the
\emph{privacy boundary} is visible in what moves: only the shared entity
embeddings, never the per-silo triples. FedV-KGQA~\cite{bappy2026fedv} does not show this per-question view, so a reader cannot see how any single question was answered.

Table~\ref{tab:trace} is one such run. For \emph{``who are the actors in the
films written by John Travis?''} the first hop \texttt{written\_by} sits in Silo
A and the second hop \texttt{starred\_actors} in Silo B, so the question is
unanswerable inside any one of them. Fusion still places a gold answer at rank 2
and recovers all three gold answers in the top six, while only entity embeddings ever
leave a silo. Stepping through other questions shows both where the mechanism fails and succeeds. The coverage reported in the trace distinguishes a failure at the
filtering stage from one where the answer was present but scored too low. The
averages in Section~\ref{sec:result} cannot make that distinction.\footnote{Source and prototype: \url{https://github.com/brains-group/fedv-kgqa-source.git} and
\url{https://github.com/brains-group/fedv-kgqa-prototype.git}}

\begin{table}[t]
  \centering
  \small
  \renewcommand{\arraystretch}{0.90}
  \setlength{\tabcolsep}{6pt}
  \caption{A live prototype run. \textbf{User selects:} \emph{``who are the
  actors in the films written by John Travis?''} The prototype then shows
  each stage in order. Gold answers: \texttt{Haley Bennett},
  \texttt{Chace Crawford}, \texttt{Jake Weber}.}
  \label{tab:trace}
  \begin{tabular}{@{}r@{\hskip 6pt}l@{\hskip 10pt}l@{}}
    \toprule
    \multicolumn{2}{@{}l}{\textbf{Stage}} & \textbf{Live output} \\
    \midrule
    \ding{182} & Topic entity  & \texttt{John Travis} (id 15311), seeks \emph{person} \\
    \addlinespace[2pt]
    \ding{183} & Silo A \textit{(private)} & (The Haunting of Molly Hartley, \texttt{written\_by}, John Travis) \\
               & Silo B \textit{(private)} & (The Haunting of Molly Hartley, \texttt{starred\_actors}, Chace Crawford) \\
               & Silo C \textit{(private)} & (The Haunting of Molly Hartley, \texttt{associated\_genre}, Horror) \\
    \addlinespace[2pt]
    \ding{184} & 2-hop filter  & 58 / 39{,}170 entities scored (0.1\%); 3/3 gold covered \\
    \addlinespace[2pt]
    \ding{185} & Fuse + encode & $\mathbf{h}_{\mathrm{joint}}[\text{John Travis}] \in \mathbb{R}^{768}$; BERT+MLP question vector \\
    \addlinespace[2pt]
    \ding{186} & Ranked output & Robert Duvall 0.29 $>$ \textbf{Jake Weber} 0.23\,\checkmark\; $>$ Sarah Polley 0.15 $>$ \dots \\
    \addlinespace[2pt]
    \midrule
    \multicolumn{3}{@{}l}{\textbf{Result:} best rank 2 \;|\; MRR 0.50 \;|\; Hits@10 \checkmark \;\; \emph{(only entity embeddings left each silo)}} \\
    \bottomrule
  \end{tabular}
\end{table}
\section{Cross-Experiment Lessons and Limitations}
\label{sec:lessons}

The results support four design lessons. Concatenation should preserve each
silo's geometric view, since averaging removes it (Finding 1). Silo count and
relation composition both matter, as moderate partitioning may aid specialization
while heavy fragmentation costs accuracy (Finding 2). Questions should be
anchored to the topic entity, since language semantics alone do not identify the
correct neighborhood (Finding 3). And communication should be reported against a
target, since the cheapest encoder depends on the operating point (Finding 3).
These lessons hold under the evaluation's assumptions of a static graph, a shared
and aligned entity vocabulary, and semantic relation partitions. Entity
embeddings reach the server, so the boundary is structural, not a formal privacy
guarantee. The candidate sets also leak information: building
$\mathcal{C}(e_0)$ across silos shows the server that an entity sits within two
hops of the topic entity, even though the relation types stay hidden. How much
a server could recover from the embeddings and the candidate sets is an open
question. Future work should study incremental updates, uncertain entity
alignment~\cite{chen2024unaligned}, alternative partitioning, message-passing encoders under a vertical split, secure aggregation,
and differential privacy. Silos may also be incomplete, since a missing edge can
break a cross-silo path that no single silo can repair locally.

\section{Conclusion}
FedV-KGQA shows that multi-hop KGQA stays feasible when relation types are split
vertically across organizations, approaching centralized accuracy, extending to
three-hop reasoning, and holding up under moderate embedding noise. Comparing the experiments directly, rather than one at a time, is what surfaces
the design lessons, and the prototype makes the mechanism inspectable one
question at a time. These findings carry one message: effective federation
takes more than keeping triples local, since enrichment, geometry-preserving
fusion, topic anchoring, and cost-aware model selection all decide whether it
works. The vertical split changes what each silo can learn, not only where
the data sits, and that is what these mechanisms have to compensate for. Whether
the same holds when graphs change over time, or when entity alignment is
imperfect, remains an important direction for future evaluation.

\section*{Declaration on Generative AI}

During the preparation of this work, the authors used Claude (Anthropic) and OpenAI Codex for
language editing, grammar correction, and structural refinement. The authors
reviewed and edited all content and take full responsibility for the submission.
No generative AI tool was used to generate research findings, experimental
results, tables, or citations. This disclosure complies with CEUR's Policy on
AI-Assisting Tools.

\bibliography{reference}

\end{document}